# A new neighborhood structure for job shop scheduling problems

Jin Xie, Xinyu Li, Liang Gao*, Lin Gui

*State Key Laboratory of Digital Manufacturing Equipment and Technology, Huazhong University of Science and Technology, Wuhan 430074, China*

**Abstract:** Job shop scheduling problem (JSP) is a widely studied NP-complete combinatorial optimization problem. Neighborhood structures play a critical role in solving JSP. At present, there are three state-of-the-art neighborhood structures, i.e., N5, N6, and N7. Improving the upper bounds of some famous benchmarks is inseparable from the role of these neighborhood structures. However, these existing neighborhood structures only consider the movement of critical operations within a critical block. According to our experiments, it is also possible to improve the makespan of a scheduling scheme by moving a critical operation outside its critical block. According to the above finding, this paper proposes a new N8 neighborhood structure considering the movement of critical operations within a critical block and the movement of critical operations outside the critical block. Besides, a neighborhood clipping method is designed to avoid invalid movement, reducing the computational time. Tabu search (TS) is a commonly used algorithm framework combined with neighborhood structures. This paper uses this framework to compare the N8 neighborhood structure with N5, N6, and N7 neighborhood structures on four famous benchmarks. The experimental results verify that the N8 neighborhood structure is more effective and efficient in solving JSP than the other state-of-the-art neighborhood structures.

## 1 Introduction

Production scheduling is critical to the planning and management of modern manufacturing enterprises. A reasonable scheduling scheme can effectively improve productivity and resource utilization. There are several styles of workshop scheduling, such as flow shop scheduling (Fernandez-Viagas and Framinan, 2019; Gmys et al., 2020; Rossit et al., 2019), job shop scheduling (Ahmadian et al., 2020; Dao et al., 2018), and flexible job shop scheduling (Gong et al., 2020a; Shen et al., 2018; Zhang et al., 2020).

Job shop scheduling problem (JSP) is a famous NP-complete problem (Lenstra et al., 1977). Given a machine set $M = \{1, 2, …, m\}$ and a job set $J = \{1, 2, …, n\}$, the objective of JSP is to determine the processing order of operations on machines for minimizing the makespan. Every job $j \in J$ is composed of $n_j$ ordered operations $O_{j1}, O_{j2}, …, O_{jn_j}$, each of them has to be processed on a predetermined machine. $O = \{0, 1, …, \tilde{n}\}$ represents the operation set, where 0 is the dummy start operation, and $\tilde{n}$ is the dummy finish operation. Every machine can only process at most one operation at a time. The processing time $P_o$ of every operation $o \in O$ is fixed. Once the processing of an operation starts, it cannot be interrupted until it is completed. There are precedence constraints between operations. Every operation $o \in O$ can be processed only if its predecessor operation has finished. Besides, every operation $o \in O$ can only be scheduled when its processing machine is idle.

We represent $S_o$ as the start time of operation $o \in O$, represent $p_o$ as its predecessor operation, and denote $E_h$ as the set of operations processed on machine $h \in M$. The model of JSP is given below:

$$\text{Minimize } C_{\max} = \max_{o \in O}\{S_o + P_o\} \tag{1}$$

Subject to:

$$S_o - S_{p_o} \geq P_{p_o};\ o = 0, 1, ..., n \qquad (2)$$

$$S_i - S_j \geq P_i\ or\ S_j - S_i \geq P_j;\ (i, j) \in E_h, h \in M \qquad (3)$$

$$S_o \geq 0;\ o = 0, 1, ..., n \qquad (4)$$

The objective function (1) minimizes the makespan. Constraint (2) guarantees the precedence order between operations of the same job. Constraint (3) ensures the precedence order between operations on the same machine. Constraint (4) imposes the start time of all operations are non-negative.

In recent years, many researchers have focused on JSP. Exact methods, including Lagrangian relaxation, linear programming, and branch and bound methods, have successfully solved small scale of instances. However, with the expansion of the scale of problems, the computational time of the exact algorithm increases exponentially. The high computational time is unacceptable. Heuristics, such as some dispatching rules, can quickly obtain a feasible solution for a large-scale instance, but it is often much worse than the optimal solution. Therefore, meta-heuristics have become more and more popular for solving JSP in recent years. Making individuals evolve in a good direction, the neighborhood structures play the very critical role in meta-heuristics. For example, Nowicki and Smutnicki (1996) first presented a fast taboo search algorithm using N5 neighborhood structure to solve JSP. Then, N5 neighborhood structure was widely applied to various algorithm frameworks. Balas and Vazacopoulos (1998) developed a guided local search procedure to solve JSP. N6 neighborhood structure was designed in this procedure to search for more feasible areas. Afterward, Zhang et al. (2007) extended N6 neighborhood structure and derived N7 neighborhood structure. N7 neighborhood structure further expanded the search space of feasible solutions compared with N5 and N6 neighborhood structure. Peng et al. (2015) presented a TS/PR algorithm that embedded TS into path relinking framework for JSP. They used N7 neighborhood structure to enhance the exploitation of their method. Cheng et al. (2016) developed a hybrid evolutionary algorithm employing N7 neighborhood structure for JSP. Besides, there are many algorithms that have achieved good results by using neighborhood structures of N5 (Abedi et al., 2020; Goncalves and Resende, 2014; Tang et al., 2019), N6 (Kurdi, 2017; Nagata and Ono, 2018; Qin et al., 2019), or N7 (Caldeira and Gnanavelbabu, 2019; Gong et al., 2020b).

Base on the above related work, it is evident that neighborhood structures are critical for solving JSP. The updating of upper bounds of the famous benchmarks is inseparable from these efficient neighborhood structures. Hence, this paper focuses on proposing an efficient neighborhood structure for JSP. N5, N6, and N7 are the most important among all the existing neighborhood structures. Observing these neighborhood structures, we find that they only consider the movement of critical operations within the critical block. However, according to our experiments, it is also possible to reduce the makespan of a scheduling scheme by moving the critical operation outside its critical block. A new neighborhood structure called as N8 is designed based on this finding. Considering the size of the N8 neighborhood structure is too large, this paper further proposes a neighborhood clipping method to avoid invalid insertion. Finally, the N8 neighborhood structure is tested on four sets of standard benchmark instances. The experimental results verify that N8 neighborhood structure is more effective and efficient in solving JSP than the other classical neighborhood structures.

The remainder of this paper is organized as follows. The disjunctive graph model of JSP is described in Section 2. A new neighborhood structure and a neighborhood clipping method are introduced in Section 3. Sections 4 presents an algorithm framework of TS. Section 5 compares the new neighborhood structure with three classical neighborhood structures, and discusses its effectiveness and superiority. Finally, Section 6 concludes this paper and gives future researches.

## 2 Disjunctive graph model for JSP

This section briefly introduces a disjunctive graph model for JSP proposed by Balas (1969) to clearly describe the N8 neighborhood structure. The disjunctive graph $G = (N, A, E)$ is used to represent any scheduling scheme, where $N$ represents the operation set, $N = \{0, O_{ij}, \tilde{n} \mid i = 1, 2, \ldots, n; j = 1, 2, \ldots, m\}$, $O_{ij}$ denotes the $j$th operation of the $i$th job, 0 is the dummy start operation, and $\tilde{n}$ is the dummy finish operation. $A$ denotes the set of conjunctive arcs that connect operations of the same job. $E$ represents the set of disjunctive arcs that connect operations processed on the same machine. $p_{ij}$ denotes the the processing time of $O_{ij}$. The length of an arc $(O_{ij}, O_{ij'}) \in A$ is equal to $p_{ij}$. The length of an arc $(O_{ij}, O_{i'j'}) \in E$ is equal to $p_{ij}$ or $p_{i'j'}$, which is depended on its orientation. Besides, the length of an arc is equal to 0, when its starting operation is 0 or ending operation is $\tilde{n}$. The longest path from 0 to $\tilde{n}$ is the critical path, its length is equal to the makespan of the scheduling scheme. The objective of obtaining the optimal scheduling scheme corresponds to finding the shortest critical path from the disjunctive graph.

Table 1

An example of JSP with three jobs and three machines

| Job | (Machine sequence, Processing time) | | |
|---|---|---|---|
| $J_1$ | (1, 2) | (2, 1) | (3, 3) |
| $J_2$ | (1, 1) | (3, 2) | (2, 2) |
| $J_3$ | (2, 5) | (1, 2) | (3, 1) |

An example of JSP with three jobs and three machines is shown in Table 1. Figure 1 gives a disjunctive graph $G = (N, A, E)$ to represent the scheduling model. Solid lines denote the conjunctive arc in $A$, and dashed lines represent the disjunctive arc in $E$. In the model, we can get

$$N = \{0, O_{11}, O_{12}, O_{13}, O_{21}, O_{22}, O_{23}, O_{31}, O_{32}, O_{33}, n\}$$

$$A = \begin{Bmatrix} (0, O_{11}), (O_{11}, O_{12}), (O_{12}, O_{13}), (O_{13}, 1) \\ (0, O_{21}), (O_{21}, O_{22}), (O_{22}, O_{23}), (O_{23}, 1) \\ (0, O_{31}), (O_{31}, O_{32}), (O_{32}, O_{33}), (O_{33}, 1) \end{Bmatrix}$$

$$E = \{E_1, E_2, E_3\}$$

$$E_1 = \{(O_{21}, O_{11}), (O_{11}, O_{21}), (O_{11}, O_{32}), (O_{32}, O_{11}), (O_{32}, O_{21}), (O_{21}, O_{32})\}$$

$$E_2 = \{(O_{31}, O_{12}), (O_{12}, O_{31}), (O_{12}, O_{23}), (O_{23}, O_{12}), (O_{23}, O_{31}), (O_{31}, O_{23})\}$$

$$E_3 = \{(O_{22}, O_{13}), (O_{13}, O_{22}), (O_{13}, O_{33}), (O_{33}, O_{13}), (O_{33}, O_{22}), (O_{22}, O_{33})\}$$

If we modify the disjunctive arc in $E_i$ ($i = 1, 2, 3$) and find a path through all nodes on each $E_i$ without circle generation, we can get a scheduling scheme as shown in Figure 2. $E_i$' denotes the new set of disjunctive arcs. $E_1$' = {$(O_{21}, O_{11})$, $(O_{11}, O_{32})$}, $E_2$' = {$(O_{31}, O_{12})$, $(O_{12}, O_{23})$}, $E_3$' = {$(O_{22}, O_{13})$, $(O_{13}, O_{33})$}. The critical path is the key element of a scheduling scheme. If $L(u, v)$ represents the length of the longest path from $u$ to $v$ in a disjunctive graph, $L(0, \tilde{n})$ denotes the length of the critical path. In Figure 2, (0, $O_{31}$, $O_{12}$, $O_{13}$, $O_{33}$, $\tilde{n}$) constitutes the critical path and its length is 10. The critical operation is defined as the operation on the critical path. The critical block consists of the adjacent critical operations on the same machine. Multiple critical blocks constitute the critical path. In Figure 2, two critical block $B_1$ = {$O_{31}$, $O_{12}$} and $B_2$ = {$O_{13}$, $O_{33}$} divides the critical path. Any operation $u$ is associated with two immediate successors and predecessors. $JS[u]$ and $JP[u]$ denote its job successor and predecessor, $MS[u]$ and $MP[u]$ denote its machine successor and predecessor. The research focus of this paper is to design a new neighborhood structure according to the critical path, i.e., modifying the critical path to try to reduce the makespan.

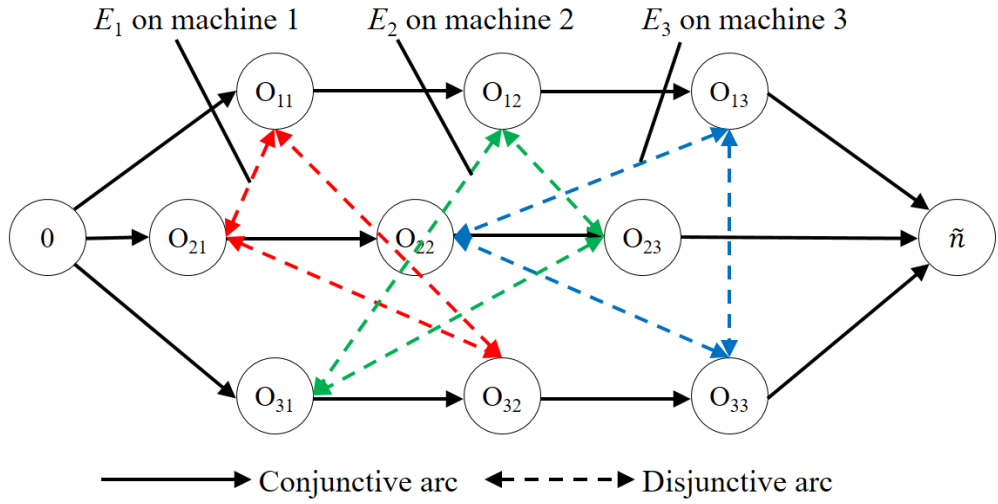

Figure 1: The disjunctive graph of an example with three jobs and three machines

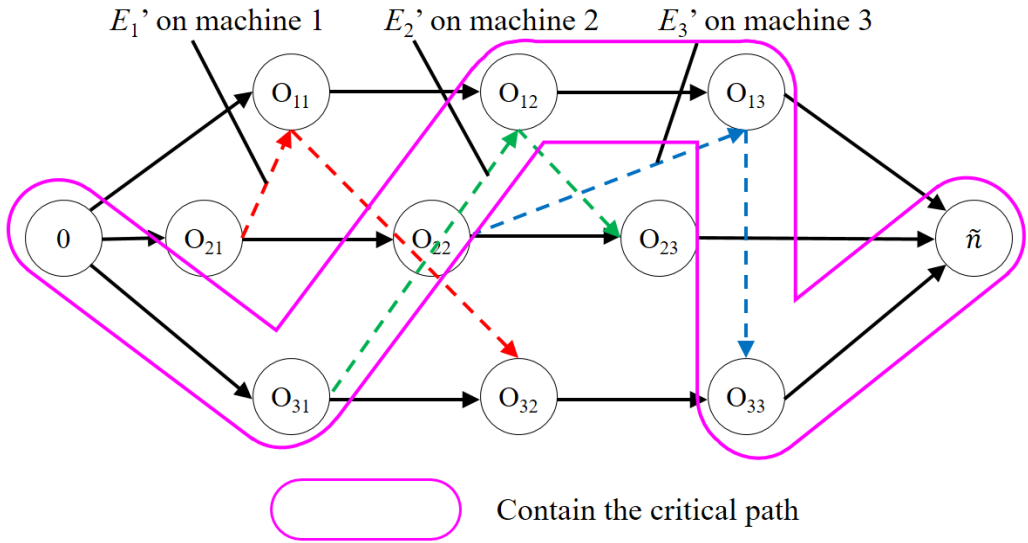

Figure 2: A feasible scheduling scheme for the example

## 3 Proposed new neighborhood structure for JSP

### 3.1 Analysis of the existing neighborhood structures

The three most famous neighborhood structures were proposed by Nowicki and Smutnicki (1996) (N5), Balas and Vazacopoulos (1998) (N6), and Zhang et al. (2007) (N7), respectively. They were designed on critical blocks. A move is defined by moving an inner operation to either the front or the rear of a critical block, or inserting either the first or the last operation inside the critical block. N5 neighborhood structure is to reverse the two front and the two rear operations in the critical block. Because the N5 neighborhood structure generates only two neighbors in a block, it has the minimum size of neighborhoods among the three neighborhood structures. N6 neighborhood structure is an extension of N5 neighborhood structure, inserting the inner operations at both ends of the critical block. N7 neighborhood structure expands N6 neighborhood structure, further moving the first and the last operation inside the critical block. N5, N6, and N7 neighborhood structures are shown in Figures 3-5.

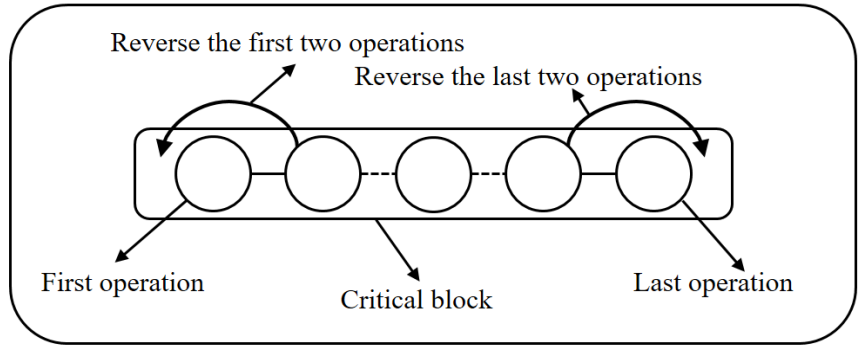

Figure 3: N5 neighborhood structure

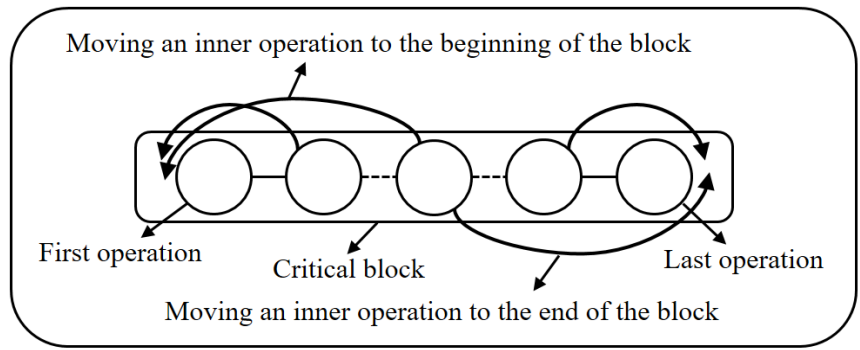

Figure 4: N6 neighborhood structure

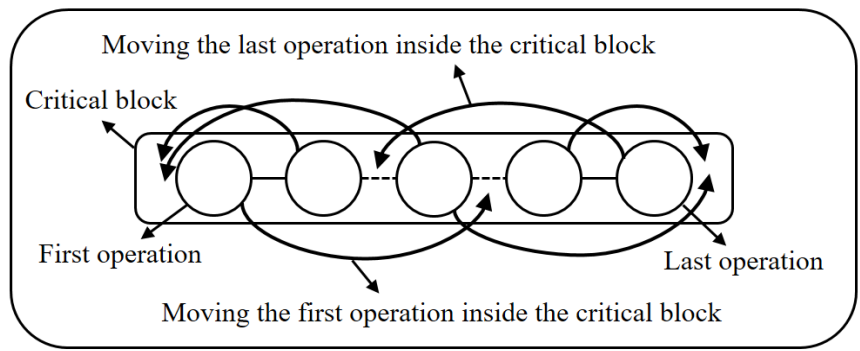

Figure 5: N7 neighborhood structure

By analyzing the above literatures on the three classic neighborhood structures, we conclude that reasonable expansion of the neighborhood structure based on critical blocks can effectively improve the global search ability of algorithms and better avoid a single individual falling into the local optimum. However, the existing neighborhood structures only consider the movement of operations within the critical block. Can the makespan of a scheduling scheme be reduced by moving the critical operation outside its critical block? In Figure 6, we give a Gantt chart of a scheduling scheme for the LA01 instance (Lawrence, 1984), and use black lines to identify its critical path. We can observe that $O_{105}$ is inside a critical block, and $O_{64}$ is outside the critical block. The makespan of the original scheduling scheme is 780. After moving $O_{105}$ right after $O_{64}$, the makespan of the new scheduling scheme is reduced to 759.

Figure 7 shows the Gantt chart of the new scheduling scheme. Similarly, Figure 6 shows $O_{95}$ is inside a critical block, and $O_{53}$ is outside the critical block. After moving $O_{95}$ right before $O_{53}$, the makespan of the new scheduling scheme is reduced to 759. Figure 8 presents the Gantt chart of the new scheduling scheme. From Figures 6-8, we can conclude that it is possible to reduce the makespan of a scheduling scheme by moving a critical operation outside its critical block. Based on this conclusion, we will introduce our new neighborhood structure in the next section.

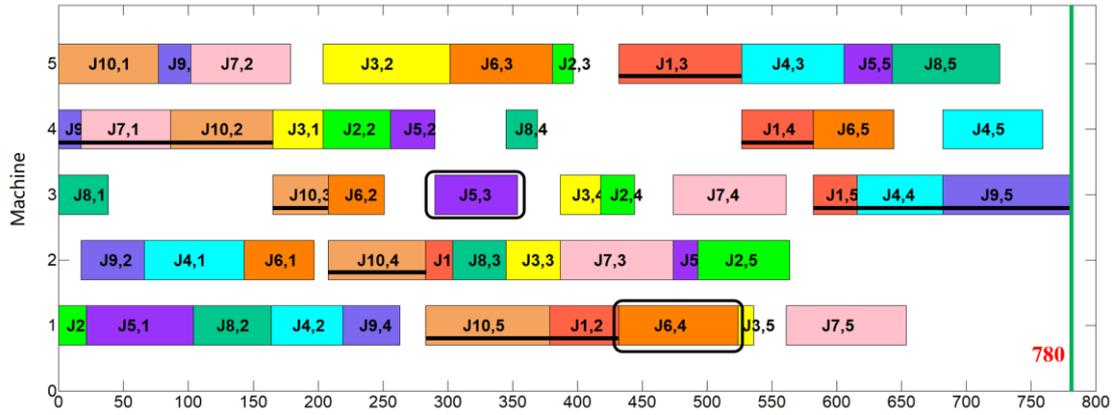

Figure 6: Gantt chart of a scheduling scheme for LA01 instance

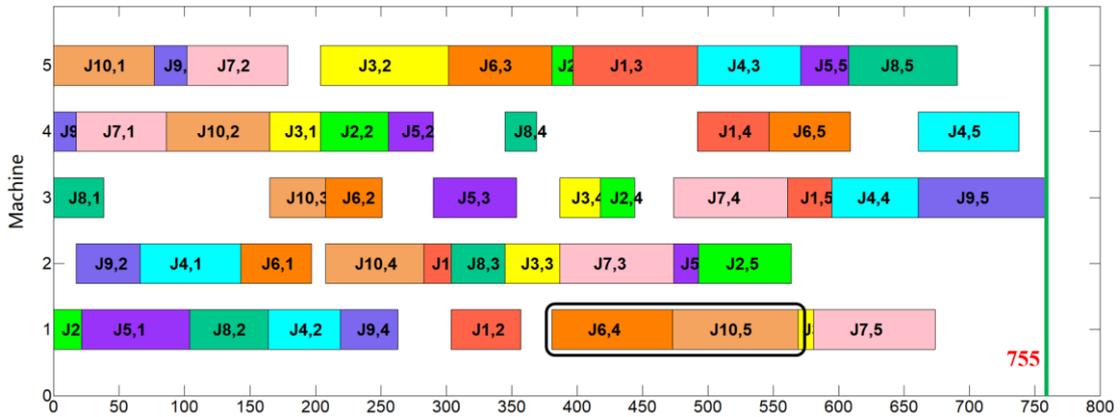

Figure 7: Gantt chart of the new scheduling scheme after moving $O_{105}$ right after $O_{64}$

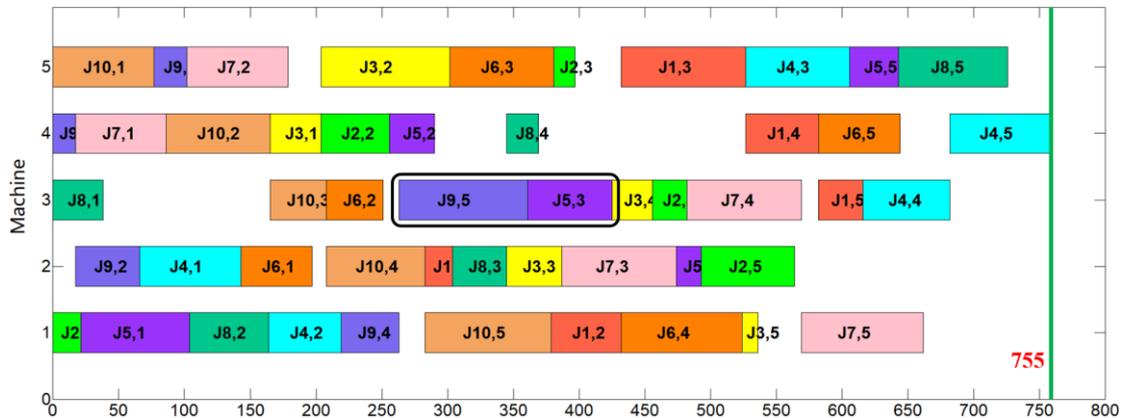

Figure 8: Gantt chart of the new scheduling scheme after moving $O_{95}$ right before $O_{53}$

## 3.2 N8 neighborhood structure

According to the analysis in Section 3.1, we find that the existing structures do not consider moving a critical operation outside its critical block. However, it is possible to reduce the makespan of a scheduling scheme by doing so. Based on this, we proposes a new neighborhood structure called N8 that extends N7. The new neighborhood structure further considers moving an inner operation (which is in a critical block, but not the first or the last one) outside its critical block, inserting the first operation after operations after the last one, and inserting the last operation before operations before the first one. The new neighborhood structure is illustrated in Figure 9. Compared with N5, N6, and N7 neighborhood structures, the N8 neighborhood structure searches for a wider solution space, increasing the probability of algorithms obtaining the optimal solution for JSP. Nevertheless, we need to ensure the feasibility of the scheduling scheme after operations are moved. Two propositions and proofs for our proposed neighborhood structure are given to ensure the generation of feasible scheduling schemes, that is no cycle is created in the disjunctive graph after operations are moved.

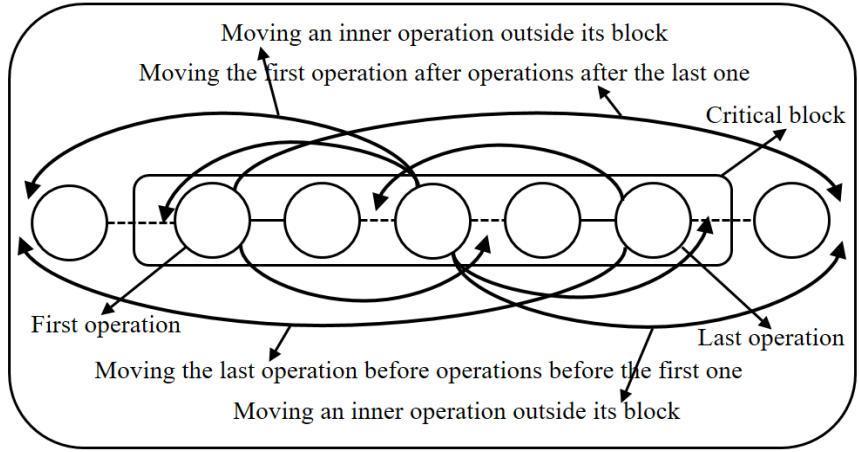

Figure 9: N8 neighborhood structure

**Proposition 1.** If operations $u$ and $v$ processed on the same machine, and $L(v, \tilde{n}) \geq L(JS[u], \tilde{n}) - p_{JS[u]}$, then moving $u$ right after $v$ generates an acyclic complete selection.

**Proof.** By contradiction: suppose a cycle $C$ is created by moving $u$ right after $v$. Thus, either $(u, MS[v])$ or $(u, JS[u])$ is contained by $C$. If $(u, MS[v]) \in C$, we can find a path from $MS[v]$ to $v$ in the disjunctive graph (the cycle is $MS[v] \rightarrow v \rightarrow u \rightarrow MS[v]$), against the assumption that the disjunctive graph is acyclic. If $(u, JS[u]) \in C$, we can find a path from $JS[u]$ to $v$ in the disjunctive graph (the cycle is $JS[u] \rightarrow v \rightarrow u \rightarrow JS[u]$), so $L(v, \tilde{n}) < L(JS[u], \tilde{n}) - p_{JS[u]}$, against the assumption. This proof is completed.

By analogy, we have Proposition 2.

**Proposition 2.** If operations $v$ and $u$ processed on the same machine, and $L(0, u) + p_u \geq L(0, JP[v])$, then moving $v$ right before $u$ generates an acyclic complete selection.

**Proof.** Parallels that of Proposition 1.

We extend Theorems 1 and 2 proposed by Zhang et al. (2007), and present Propositions 1 and 2 to construct N8 neighborhood structure. Propositions 1 and 2 relax the inequality constraints of Theorems 1 and 2, thereby expanding the size of neighborhood solutions. Besides, Theorems 1 and 2 are designed for operations in a critical block. However, our proposed Proposition 1 and 2 can be applied to any operations whether they are in a critical block or not. Because only by modifying the critical path can the makespan of a scheduling scheme be reduced, we ensure that at least one of the two operations $u$ and $v$ must be in a critical block for the N8 neighborhood structure. Thus, the neighbors of a feasible solution can be obtained by performing an insertion for operations *u* and *v* as follows: 1) if both *u* and *v* are in the same critical block, the insertion is the same as that of the N7 neighborhood structure; 2) if *u* is an operation in a critical block, *v* is an operation after the block, and Proposition 1 is satisfied, then insert *u* right after *v*; 3) if *v* is an operation in a critical block, *u* is an operation before the block, and Proposition 2 is satisfied, then insert *v* right before *u*. All of these insertion operators are shown in Figure 9. Our neighborhood structure can obtain the most neighbors compared to those obtained by N5, N6, and N7, but an insertion for operations *u* and *v* may result in the makespan increase. How to avoid invalid insertion has become another focus of our research. The method of neighborhood clipping will be introduced in the next section. It can effectively reduce the size of neighbor solutions.

## 3.3 Neighborhood clipping

The search space of the N8 neighborhood structure is much larger than that of N5, N6, and N7 neighborhood structures. However, too many neighborhood solutions increase the estimation times, thereby increasing the computational time of algorithms. Hence, we propose a neighborhood clipping method, including four propositions and proofs, to avoid invalid insertion and reduce the size of neighbor solutions.

**Proposition 3.** If *u* is the first operation and *v* is the inner operation (which is not the first or last operation) in the first critical block, then moving *u* right after *v* cannot reduce the makespan.

**Proof.** The makespan of the original scheduling scheme is $L(0, \tilde{n}) = p_u + L(MS[u], v) + p_v + L(MS[v], \tilde{n})$. After moving operation *u* right after *v*, the makespan of the new scheduling scheme is $L'(0, \tilde{n}) = L(0, MS[u]) + L(MS[u], v) + p_v + p_u + L(MS[v], \tilde{n})$. Because $L(0, MS[u]) \geq 0$, $L'(0, \tilde{n}) \geq L(0, \tilde{n})$. Hence, moving *u* right after *v* does not reduce the makespan. The proof is completed. Figure 10 shows the illustration of moving operation *u* right after *v* in the first block.

By analogy, we have Proposition 4.

**Proposition 4.** If *v* is the last operation and *u* is the inner operation in the last critical block, then moving *v* right before *u* cannot reduce the makespan.

**Proof.** Parallels that of Proposition 3.

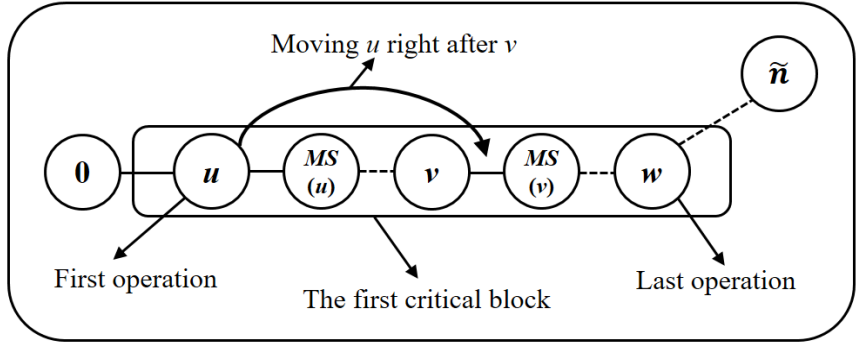

(a) Before move

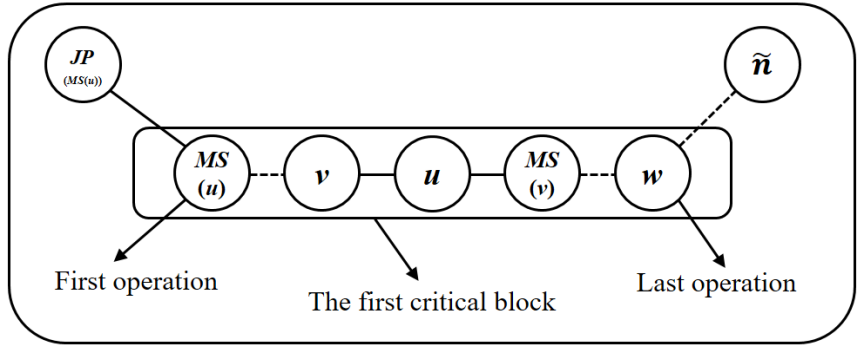

(b) After move

Figure 10: Illustration of moving operation *u* right after *v* in the first block

**Proposition 5.** If *v* is the inner operation and *u* is the first operation in the first critical block, then moving *v* right before *u* cannot reduce the makespan.

**Proof.** The makespan of the original scheduling scheme is $L(0, \tilde{n}) = p_u + L(MS[u], v) + p_v + L(MS[v], \tilde{n})$. After moving *v* right before *u*, the makespan of the new scheduling scheme is $L'(0, \tilde{n}) = L(0, v) + p_v + p_u + L(MS[u], v) + L(MS[v], \tilde{n})$. Because $L(0, v) \geq 0$, $L'(0, \tilde{n}) \geq L(0, \tilde{n})$. Hence, moving *v* right before *u* does not reduce the makespan. The proof is completed. Figure 11 shows the illustration of moving operation *v* right before *u* in the first block.

By analogy, we have Proposition 6.

**Proposition 6.** If *u* is the inner operation and *v* is the last operation in the last critical block, then moving *u* right after *v* cannot reduce the makespan.

**Proof.** Parallels that of Proposition 5.

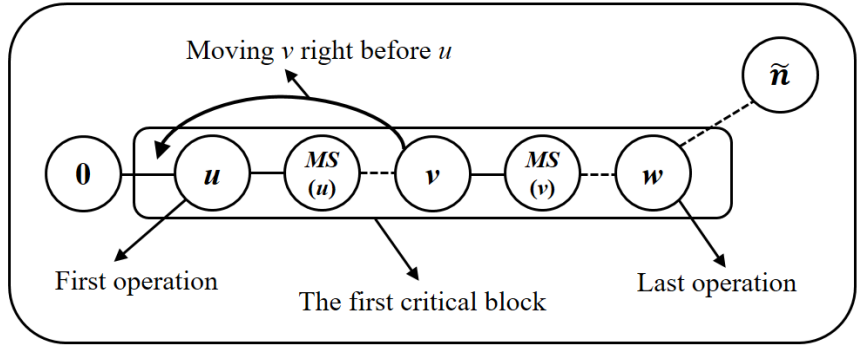

(a) Before move

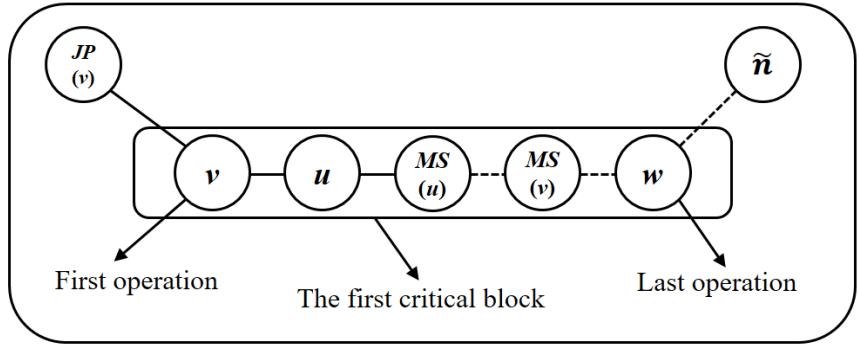

(b) After move

Figure 11: Illustration of moving operation *v* right before *u* in the first block

Our proposed four propositions can effectively avoid invalid insertion and reduce the size of neighbor solutions. Specifically, Proposition 3 and 5 are applied to the first critical block. Proposition 3 avoids the invalid insertion of moving the first operation after any inner operator. Proposition 5 avoids the invalid insertion of moving any inner operation before the first operator. Proposition 4 and 6 are used to the last critical block. Proposition 4 avoids the invalid insertion of moving the last operation before any inner operator. Proposition 6 avoids the invalid insertion of moving any inner operation after the last operator. After describing the new neighborhood structure and the neighborhood clipping method, we will introduce the algorithm framework of TS-N8 in the next section.

## 4 TS algorithm with N8 neighborhood structure for JSP

### 4.1 Main framework

Tabu search (TS) algorithm is an effective method for solving shop scheduling problems. The characteristic of the algorithm is to use a tabu list to avoid itself getting stuck at the local optimum. If a solution is forbidden in the tabu list, it cannot be selected as a new solution for the next generation.

The procedure of TS-N8 is designed as follows: 1) randomly initialize a candidate solution $x_0$ and take it as the current optimal solution $x_b$; 2) set the tabu list $T$ to be empty; 3) set $N$ to 0 and $N_{max}$ to *improveIter*. $N$ denotes the number of generations for $x_b$, which has not been improved. If $N$ reaches the maximum number of unimproved generations $N_{max}$, randomly choose a neighborhood solution of $x_0$ to update it; 4) based on N8 neighborhood structure, generate the set of neighborhood solutions *Children* of $x_0$; 5) pick the optimal solution *x'* which is not forbidden or satisfies aspiration criterion from *Children*; 6) if the fitness of *x'* better than that of $x_b$, set $x_b \leftarrow x'$ and $N \leftarrow 0$, else set $N \leftarrow N + 1$; 7) If $N == N_{max}$, randomly select a solution *x"* from *Children*, set $x_0 \leftarrow x"$ and $N \leftarrow 0$, else $x_0 \leftarrow x'$; 8) update the tabu list $T$; 9) if the stop criterion (i.e., the makespan of $x_b$ reaches the lower bound or the total iteration number achieves a maximum value) is satisfied, output the optimal solution $x_b$. The pseudo-code of TS-N8 is illustrated in Figure 12.

| TS-N8 | |
|---|---|
| 1 | Randomly initialize a candidate solution $x_0$ |
| 2 | $x_b \leftarrow x_0$ |
| 3 | $T \leftarrow \emptyset$ // $T$ is the tabu list |
| 4 | $N \leftarrow 0$, $N_{max} \leftarrow$ *improveIter* |
| 5 | **While** (stop criterion is not satisfied) **do** |
| 6 |     *Children* $\leftarrow \emptyset$ // *Children* is the set of neighborhood solutions of $x_0$ |
| 7 |     **While** (\|*Children*\| < M) // M is the size of neighborhood solutions **do** |
| 8 |         generate a neighborhood solution *x'* of $x_0$ |
| 9 |         *Children* $\leftarrow$ *Children* $\cup$ *x'* |
| 10 |     **End while** |
| 11 |     *x'* $\leftarrow$ arg min ($f(x) : x \in$ *Children* $\wedge$ ($x$ is not tabu $\vee$ $x$ satisfies aspiration criterion)) |
| 12 |     **If** $f(x') < f(x_b)$ **then** |
| 13 |         $x_b \leftarrow x'$, $N \leftarrow 0$ |
| 14 |     **Else** |
| 15 |         $N \leftarrow N + 1$ |
| 16 |     **End if** |

| 17 | **If** $N == N_{\max}$ **then** |
| --- | --- |
| 18 | Randomly select a solution $x''$ from *Children* |
| 19 | $x_0 \leftarrow x''$, $N \leftarrow 0$ |
| 20 | **Else** |
| 21 | $x_0 \leftarrow x'$ |
| 22 | **End if** |
| 23 | Update $T$ |
| 24 | **End while** |
| 25 | **Return** the optimal solution $x_b$ |

Figure 12. The pseudo-code of the TS-N8 algorithm

**4.2 Move evaluation**

After an insertion operator, the makespan of a new scheduling scheme needs to be recalculated. Ten et al.(1999) proposed an exact estimation method to calculate the makespan of the new scheduling scheme. The core idea of their approach is only to recalculate the starting and the finish time of an operation affected by the insertion operator. If an operation is not affected by the insertion operation, its start and end time are not recalculated. However, the exact estimation method still consumes a lot of computational time for large-scale problems. To reduce the computational time, Taillard (1994) first proposed a simple and easy-to-use fast estimation method for JSP. Then Nowicki and Smutnicki (2002) further developed Taillard's method and presented a more accurate fast estimation method for JSP. Nevertheless, the above estimation methods are only applicable for swapping adjacent operations in a critical block. Balas and Vazacopoulos (1998) presented a more efficient fast estimation method for N6 neighborhood structure to estimate the start and the end time of an operation after multiple disjunctives are reversed. Because this method is adapted to reverse multiple disjunctives, it also can effectively be applied to our neighborhood structure.

**4.3 Tabu list and tabu tenure**

TS uses the tabu list to avoid re-visiting the solutions traversed in the previous steps. For N7 neighborhood structure, the solution attributes instead of move attributes are forbidden in the tabu list. Similar to N7 neighborhood structure, our neighborhood structure also needs to consider the reversal of multiple disjunctives after an insertion operator. TS-N8 uses the same tabu list designed for the N7 neighborhood structure. Specifically, if a move is composed of a swap on $u$ and $v$ and achieves the same operation sequence and positions on machines from $u$ to $v$, it is forbidden for the duration. A disadvantage of the tabu list is that it would cause the prohibition of a good solution having not been visited. We use an aspiration criterion to avoid the problem. The criterion adopts the move if the new solution is better than the current optimal solution.

Tabu tenure determines the time limit of moves that are forbidden at the current iteration. If it is too short, the algorithm cannot avoid the circle generation. On the contrary, if it is too long, many unvisited solutions will be missed. Hence, the length of tabu tenure is a critical parameter for TS-N8. Literature (Zhang et al., 2007) shows the appropriate length of the tabu list increases with the increase of the ratio of the number of jobs ($n$) to the number of machines ($m$). When the minimum length of the tabu list is set to $L = 10 + n / m$, TS-N8 can obtain good solutions. Furthermore, we use a dynamic tabu list to avoid TS-N8 falling into cycling. The length of the dynamic tabu list is chosen in the range of [$L_{\min}$, $L_{\max}$] randomly, where $L_{min} = L$ and $L_{max} = 1.5L$.

### 4.4 Move selection

The procedure of TS-N8 chooses the move which is with the lowest makespan and not be forbidden or meets the aspiration criterion in each iteration. Nevertheless, it may happen that all candidate moves are forbidden, and the aspiration criterion are also not met. Under this circumstance, a move is randomly chosen from all possible moves. Besides, if the current optimal solution has not been improved within a predetermined iteration number (*ImproveIter*), a move is randomly chosen from all possible moves, too. According to a significant number of experiments, we found that setting *ImproveIter* to 200 could get good results.

## 5 Computational results

### 5.1 Experimental settings

This section uses 110 well-known benchmark instances to test the effectiveness of N8 neighborhood structure. These job shop scheduling instances are as follows: an instance denoted as FT10 due to Fisher and Thompson (1963), 40 instances LA01-40 due to Lawrence (1984), 3 instances ABZ7-9 due to Adams et al. (1988), 10 instances ORB01-10 due to Applegate and Cook (1991), 6 instances SWV10-15 due to Storer et al. (1992) and 50 instances TA01-50 due to Taillard (1993).These benchmark instances contain small, medium and large scale JSPs, which are very difficult to solve. For example, the upper and the lower bounds of some instances on ABZ and TA benchmark are still not equal. In order to eliminate the influence of randomness, all instances are tested independently over ten runs.

The three most famous algorithms, TSAB-N5 (Nowicki and Smutnicki, 1996), SBGLS-N6 (Balas and Vazacopoulos, 1998), and TS-N7 (Zhang et al., 2007), are selected as comparison algorithms. TSAB-N5 is a TS algorithm employing the N5 neighborhood structure. It works fast and efficiently, and has solved a well-known $10 \times 10$ hard benchmark problem within only 30 seconds. SBGLS-N6 is a guided local search procedure, which can effectively overcome local optima by employing the N6 neighborhood structure. It has found most of the optimal solutions on LA, TA and ORB instances. TS-N7 is a TS algorithm using N7 neighborhood structure. It has found two better upper bounds on SWV instances. These three algorithms have shown the superiority of their neighborhood structure. We compare TS-N8 with them to demonstrate the effectiveness of our proposed neighborhood structure.

The mean relative error (MRE) is used to measure the quality of solutions. It can be calculated by the relative deviation equation RE = $100 \times (UB_{solve} - LB_{best}) / LB_{best}$, where $LB_{best}$ is the lower bound, and $UB_{solve}$ is the best-known makespan.

The initial solution is randomly generated in TS-N8. Section 4.3 gives the length of tabu tenure. The parameter of *ImproveIter* is set to 200, and the maximum iteration number is set to 50 million. Our algorithm was implemented in Java and ececuted on an Intel Core 3.6-GHz PC with 16-GB memory. In the remainder of this section, four sets of benchmark instances are used to measure the effectiveness of our neighborhood structure.

### 5.2 Computational results on FT, LA, and ABZ instances

Firstly, fifteen difficult instances are selected in FT, LA, and ABZ sets for performance testing. Table 1 provides the detailed comparison results between TS-N8 and TSAB-N5, SBGLS-N6 and TS-N7 for the fifteen difficult instances. The best known upper (lower) bounds are listed in column UB (LB).

The following three columns, Best, $M_{av}$, and $T_{av}$ denote the best makespan, average makespan, and average computational time of TS-N8 obtaining the optimal solution. The next three columns show the same indicators of TS-N7. The last two columns give the best results obtained by TSAB-N5 and SBGLS-N6. The last row provides the best MRE (b-MRE) and the average MRE (av-MRE) for analyzing the performance of the above methods.

We can observe that the values of b-MRE obtained by TS-N8 is lower than those of other algorithms, and the values of av-MRE obtained by TS-N8 is lower than those of TS-N7 in Table 1. Moreover, TS-N8 obtains the solutions for LA29-1153, LA40-1222, and ABZ-668 better than those obtained by other algorithms do. Although the initial solutions are randomly generated, the values of $M_{av}$ are very close to those of Best obtained by TS-N8. It indicates that TS-N8 is robust in solving JSP. Besides, the values of $T_{av}$ obtained by TS-N8 are shorter than those obtained by TS-N7 for most instances, which shows that TS-N8 solves JSP more efficiently than TS-N7.

Table 2 gives a summary of comparison results between TS-N8, TSAB-N5, and SBGLS-N6 for LA instances. We do not compare TS-N8 and TS-N7 because TS-N7 did not provide complete data for each instance. For each group of instances, b-MRE and $T_{av}$ are listed for each algorithm, and TS-N8 provides the av-MRE. The results verify that TS-N8 outperforms TSAB-N5 and SBGLS-N6 with regard to solution quality and computational time. Specifically, the values of av-MRE obtained by TS-N8 are lower than those obtained by TSAB-N5 and SBGLS-N6. The average computational time of TS-N8 is shorter than that of TSAB-N5 and SBGLS-N6.

Table 1

Computational results for FT, LA, and ABZ instances

| Problem | Size | UB (LB) | TS-N8 | | | TS-N7 | | | TSAB-N5 | SBGLS-N6 |
|---|---|---|---|---|---|---|---|---|---|---|
| | | | Best | $M_{av}$ | $T_{av}$(s) | Best | $M_{av}$ | $T_{av}$(s) | | |
| FT10 | 10 ×10 | 930 | 930 | 930 | 9.8 | 930 | 930.4 | 41.1 | 930 | 930 |
| LA19 | 10 ×10 | 842 | 842 | 842 | 1.0 | 842 | 842 | 5.6 | 842 | 842 |
| LA21 | 15 ×10 | 1046 | 1046 | 1046 | 73.3 | 1046 | 1046.6 | 113.3 | 1047 | 1046 |
| LA24 | 15 ×10 | 935 | 935 | 935 | 49.1 | 935 | 937.1 | 87.5 | 939 | 935 |
| LA25 | 20 ×10 | 977 | 977 | 977 | 9.8 | 977 | 977.1 | 54.7 | 977 | 977 |
| LA27 | 20 ×10 | 1235 | 1235 | 1235 | 6.7 | 1235 | 1235 | 16.9 | 1236 | 1235 |
| LA29 | 20 ×10 | 1152 | **1153** | 1153.3 | 192.4 | 1156 | 1160.9 | 189.7 | 1160 | 1157 |
| LA36 | 15 ×15 | 1268 | 1268 | 1268 | 18.2 | 1268 | 1268.1 | 59.7 | 1268 | 1268 |
| LA37 | 15 ×15 | 1397 | 1397 | 1398.5 | 223.5 | 1397 | 1406.6 | 172.8 | 1407 | 1397 |
| LA38 | 15 ×15 | 1196 | 1196 | 1196 | 104.3 | 1196 | 1200.9 | 181.3 | 1196 | 1196 |
| LA39 | 15 ×15 | 1233 | 1233 | 1233 | 105.5 | 1233 | 1234.3 | 145.5 | 1233 | 1233 |
| LA40 | 15 ×15 | 1222 | **1222** | 1223 | 38.1 | 1224 | 1224.5 | 167.8 | 1229 | 1224 |
| ABZ7 | 20 ×15 | 656 | 657 | 657.9 | 166.4 | 657 | 661.2 | 243.4 | 670 | 662 |
| ABZ8 | 20 ×15 | 667 (648) | **668** | 668.6 | 193.1 | 669 | 670.2 | 270.9 | 682 | 669 |
| ABZ9 | 20 ×15 | 678 | **679** | 680 | 219.8 | 680 | 684.7 | 263.6 | 695 | 679 |
| MRE | - | - | **0.23** | 0.27 | - | 0.28 | 0.56 | - | 0.83 | 0.33 |

The best solutions of TS-N8 better than those of compared algorithms are indicated in bold.

Table 2

A summary of comparisons for LA instances

| Problem | size | TS-N8 | TSAB-N5 | SBGLS-N6 |
|---|---|---|---|---|

| | group | | b-MRE | av-MRE | $T_{av}$(s) | b-MRE | $T_{av}$(s) | b-MRE | $T_{av}$(s) |
|---|---|---|---|---|---|---|---|---|---|
| LA01-05 | 10 ×5 | | 0.00 | 0.00 | 0.03 | 0.00 | 3.8 | 0.00 | 3.9 |
| LA16-20 | 10 ×10 | | 0.00 | 0.00 | 0.25 | 0.02 | 68.8 | 0.00 | 25.1 |
| LA21-25 | 15 ×10 | | 0.00 | 0.00 | 26.1 | 0.10 | 74 | 0.00 | 314.6 |
| LA26-30 | 20 ×10 | | 0.02 | 0.02 | 36.8 | 0.16 | 136.4 | 0.09 | 100.0 |
| LA36-40 | 15 ×15 | | 0.00 | 0.04 | 98.0 | 0.28 | 375.6 | 0.03 | 623.5 |
| Average | - | | 0.00 | 0.01 | 32.2 | 0.11 | 131.72 | 0.02 | 213.42 |

### 5.3 Computational results on ORB instances

To further evaluate the effectiveness of our proposed neighborhood, we compare TS-N8 with TSAB-N5 and SBGLS-N6 for ORB instances, which contain ten instances with size 10 ×10. We do not compare TS-N8 and TS-N7 because TS-N7 did not provide the experimental results on this benchmark. In Section 4.4, we will introduce SWV instances to compare TS-N8 with TS-N7.

Table 3 summarizes the comparison results for the ORB instances. The results show that TS-N8 performs better than TSAB-N5 and SBGLS-N6. TS-N8 obtains the optimal solution for all instances. However, TSAB-N5 receives the optimal solution for three out of ten instances, and SBGLS-N6 receives the optimal solution for eight out of ten instances. Furthermore, the b-MRE of TS-N8 is lower than that of TSAB-N5 and SBGLS-N6. The average computational time of TS-N8 is shorter than that of TSAB-N5 and SBGLS-N6. The experimental results demonstrate that TS-N8 is more robust and efficient in solving JSP than TSAB-N5 and SBGLS-N6.

Table 3

Computational results for ORB instances

| Problem group | size | LB | TS-N8 | | | TSAB-N5 | | SBGLS-N6 | |
|---|---|---|---|---|---|---|---|---|---|
| | | | Best | $M_{av}$ | $T_{av}$(s) | Best | CI-CPU | Best | CI-CPU |
| ORB01 | 10 ×10 | 1059 | 1059 | 1059 | 3.2 | 1059 | 548 | 1059 | 17.3 |
| ORB02 | 10 ×10 | 888 | 888 | 888 | 2.5 | 890 | 376 | 888 | 88.4 |
| ORB03 | 10 ×10 | 1005 | 1005 | 1005 | 2.5 | 1005 | 356 | 1005 | 16.2 |
| ORB04 | 10 ×10 | 1005 | **1005** | 1005 | 37.6 | 1011 | 427 | 1013 | 285.6 |
| ORB05 | 10 ×10 | 887 | **887** | 887 | 7.9 | 889 | 389 | 889 | 15.2 |
| ORB06 | 10 ×10 | 1010 | 1010 | 1010 | 5.9 | 1013 | 472 | 1010 | 124.8 |
| ORB07 | 10 ×10 | 397 | 397 | 397 | 0.9 | 397 | 642 | 397 | 69.2 |
| ORB08 | 10 ×10 | 899 | 899 | 899 | 5.6 | 913 | 568 | 899 | 97.6 |
| ORB09 | 10 ×10 | 934 | 934 | 934 | 1.1 | 941 | 426 | 934 | 73.2 |
| ORB10 | 10 ×10 | 944 | 944 | 944 | 0.3 | 946 | 667 | 944 | 14.2 |
| MRE | - | - | **0.00** | 0.00 | - | 0.37 | - | 0.10 | - |

The best solutions of TS-N8 better than those of compared algorithms are indicated in bold.

### 5.4 Computational results on TA instances

There are 80 instances with the operation number between 225 and 2000 in TA benchmark. It is one of the most widely used and difficult instances over the last 20 years. Because TA51-80 are relatively easy, our experiment focuses on the hard instances TA01-50. Same as the ORB instances, TS-N7 did not give the experimental data for TA instances. Thus, we can only compare TS-N8 with TSAB-N5 and SBGLS-N6. Table 4 presents the computational results for TA instances. For 19 out of 21 instances, the best solutions obtained by TS-N8 are better than those obtained by TSAB-N5. For 35 out of 50 instances, the best solutions obtained by TS-N8 are better than those obtained by SBGLS-N6. Moreover, the b-MRE of TS-N8 is 1.28%, which is lower than 1.74% of TSAB-N5 and 1.67% of SBGLS-N6. In sum, the experimental results demonstrate that TS-N8 is competitive with TSAB-N5 and SBGLS-N6.

Table 4

Computational results for TA instances

| Problem | Size | UB (LB) | TS-N8 | | | TSAB-N5 | SB-GLS-N6 |
|---------|------|---------|-------|------|---------|---------|--------|
|         |      |         | Best  | $M_{av}$ | $T_{av}(s)$ | | |
| TA01 | 15 ×15 | 1231 | 1231 | 1231 | 19.4 | - | 1231 |
| TA02 | 15 ×15 | 1244 | 1244 | 1244 | 23.5 | 1244 | 1244 |
| TA03 | 15 ×15 | 1218 | 1218 | 1218.9 | 92.9 | 1222 | 1218 |
| TA04 | 15 ×15 | 1175 | **1175** | 1175 | 87.2 | - | 1181 |
| TA05 | 15 ×15 | 1224 | **1224** | 1224 | 43.5 | 1233 | 1233 |
| TA06 | 15 ×15 | 1238 | **1238** | 1240 | 103.7 | - | 1241 |
| TA07 | 15 ×15 | 1227 | 1228 | 1228 | 1.5 | - | 1228 |
| TA08 | 15 ×15 | 1217 | 1217 | 1217 | 24.7 | 1220 | 1217 |
| TA09 | 15 ×15 | 1274 | 1274 | 1274 | 131.0 | 1282 | 1274 |
| TA10 | 15 ×15 | 1241 | 1241 | 1242 | 90.9 | 1259 | 1241 |
| TA11 | 20 ×15 | 1357 | **1361** | 1364.4 | 209.1 | - | 1364 |
| TA12 | 20 ×15 | 1367 | 1373 | 1374.1 | 167.1 | 1377 | 1367 |
| TA13 | 20 ×15 | 1342 | **1348** | 1350.1 | 204.1 | - | 1350 |
| TA14 | 20 ×15 | 1345 | 1345 | 1345 | 54.0 | 1345 | 1345 |
| TA15 | 20 ×15 | 1339 | **1342** | 1344.2 | 214.9 | - | 1353 |
| TA16 | 20 ×15 | 1360 | **1360** | 1363.3 | 276.9 | - | 1369 |
| TA17 | 20 ×15 | 1462 | **1471** | 1473.1 | 285.0 | - | 1478 |
| TA18 | 20 ×15 | 1396 (1377) | 1398 | 1402.8 | 248.7 | 1413 | 1396 |
| TA19 | 20 ×15 | 1332 | **1335** | 1337.6 | 290.5 | 1352 | 1341 |
| TA20 | 20 ×15 | 1348 | **1352** | 1354.4 | 283.3 | 1362 | 1359 |
| TA21 | 20 ×20 | 1642 | **1644** | 1646.7 | 306.2 | - | 1659 |
| TA22 | 20 ×20 | 1600 (1561) | **1600** | 1606.2 | 291.3 | - | 1603 |
| TA23 | 20 ×20 | 1557 (1518) | 1561 | 1564.4 | 310.7 | - | 1558 |
| TA24 | 20 ×20 | 1644 | **1652** | 1653.7 | 230.7 | - | 1659 |
| TA25 | 20 ×20 | 1595 (1558) | **1598** | 1598 | 317.7 | - | 1615 |
| TA26 | 20 ×20 | 1645 (1591) | **1651** | 1653.5 | 326.3 | 1657 | 1659 |
| TA27 | 20 ×20 | 1680 (1652) | **1686** | 1687.5 | 378.9 | - | 1689 |
| TA28 | 20 ×20 | 1603 | 1615 | 1615.6 | 191.3 | - | 1615 |
| TA29 | 20 ×20 | 1625 (1583) | **1627** | 1628.5 | 308.1 | 1629 | 1629 |
| TA30 | 20 ×20 | 1584 (1528) | **1585** | 1588.3 | 253.0 | - | 1604 |
| TA31 | 30 ×15 | 1764 | **1764** | 1764.3 | 17.7 | 1766 | 1766 |
| TA32 | 30 ×15 | 1784 (1774) | 1803 | 1805.8 | 344.0 | 1841 | 1803 |
| TA33 | 30 ×15 | 1791 (1788) | 1797 | 1803.1 | 314.6 | 1832 | 1796 |
| TA34 | 30 ×15 | 1828 | **1831** | 1831.3 | 35.3 | - | 1832 |
| TA35 | 30 ×15 | 2007 | 2007 | 2007 | 0.8 | - | 2007 |
| TA36 | 30 ×15 | 1819 | **1819** | 1819 | 168.2 | - | 1823 |
| TA37 | 30 ×15 | 1771 | **1782** | 1783.4 | 357.2 | 1815 | 1784 |
| TA38 | 30 ×15 | 1673 | **1673** | 1673 | 340.4 | 1700 | 1681 |
| TA39 | 30 ×15 | 1795 | **1795** | 1795 | 352.9 | 1811 | 1798 |
| TA40 | 30 ×15 | 1670 (1651) | **1678** | 1681.3 | 410.0 | 1720 | 1686 |
| TA41 | 30 ×20 | 2006 (1906) | **2019** | 2023 | 740.4 | - | 2026 |
| TA42 | 30 ×20 | 1939 (1884) | **1954** | 1959.7 | 788.7 | - | 1967 |
| TA43 | 30 ×20 | 1846 (1809) | **1862** | 1869 | 557.4 | - | 1881 |
| TA44 | 30 ×20 | 1979 (1948) | **1989** | 1993.9 | 623.2 | - | 2004 |
| TA45 | 30 ×20 | 2000 (1997) | **2001** | 2006 | 661.2 | - | 2008 |
| TA46 | 30 ×20 | 2006 (1957) | **2022** | 2027.5 | 644.5 | - | 2040 |
| TA47 | 30 ×20 | 1889 (1807) | **1910** | 1913.8 | 654.8 | - | 1921 |
| TA48 | 30 ×20 | 1937 (1912) | **1956** | 1959.6 | 770.4 | 2001 | 1982 |
| TA49 | 30 ×20 | 1963 (1931) | **1974** | 1977.5 | 752.3 | - | 1994 |
| TA50 | 30 ×20 | 1923 (1833) | **1931** | 1936.2 | 667.7 | - | 1967 |
| MRE | - | - | **1.28** | 1.42 | - | 1.74 | 1.67 |

The best solutions of TS-N8 better than those of compared algorithms are indicated in bold.

## 5.5 Computational results on SWV instances

The last experiment is based on the SWV11-15 instances, which are widely applied to performance evaluation algorithms and challenging to be solved. Because TS-N7 did not provide the computational results for TA and ORB instances and only presented the experimental results for SWV11-15 instances, we use these instances to compare TS-N8 with TS-N7 to evaluate the effectiveness of our neighborhood further. Besides, we provide the compared results of TS-N8 with SBGLS-N6. From Table 5, it is easily observed that TS-N8 outperforms TS-N7 and SBGLS-N6 with regard to solution quality and computational time. Specially, all of the best solutions obtained by TS-N8 are better than those obtained by SBGLS-N6. The solutions for SWV12-2977 and SWV15-2885 obtained by TS-N8 are better than those obtained by TS-N7. Furthermore, the b-MRE of TS-N8 is lower than that of TS-N7 and SBGLS-N6. The values of $T_{av}$ obtained by TS-N8 are shorter than those obtained by TS-N7 for all instances. The experimental results verify that TS-N8 is more effective and efficient in solving JSP than TS-N7 and SBGLS-N6.

Table 5

Computational results for SWV11-15 instances

| Problem group | size | UB (LB) | TS-N8 | | | TS-N7 | | SBGLS-N6 |
|---|---|---|---|---|---|---|---|---|
| | | | Best | $M_{av}$ | $T_{av}$(s) | Best | $T_{av}$(s) | |
| SWV11 | 50 ×10 | 2983 | 2983 | 2983 | 601.5 | 2983 | 823 | 3008 |
| SWV12 | 50 ×10 | 2972 | **2977** | 2988.1 | 4102.3 | 2979 | 8631 | 3041 |
| SWV13 | 50 ×10 | 3104 | 3104 | 3104 | 734.2 | 3104 | 297 | 3163 |
| SWV14 | 50 ×10 | 2968 | 2968 | 2968 | 434.7 | 2968 | 168 | 2968 |
| SWV15 | 50 ×10 | 2885 | **2885** | 2891.0 | 4284.0 | 2886 | 5672 | 2942 |
| MRE | - | - | **0.03** | 0.15 | - | 0.05 | - | 1.41 |

The best solutions of TS-N8 better than those of compared algorithms are indicated in bold.

## 5.6 Discussions on the new neighborhood structure

From the experimental results of the above benchmark instances, we can clearly observe that N8 neighborhood structure achieves better solutions than the classical N5, N6 and N7 neighborhood structures. Besides, the TS algorithm using N8 neighborhood structure costs shorter computational time than the other algorithms employing N5, N6 and N7 neighborhood structures, which shows that N8 neighborhood structure has high search efficiency. Hence, these experimental results fully demonstrate that N8 neighborhood structure is effective and efficient in solving JSP.

The advantages of N8 neighborhood structure are as follows. Firstly, N8 neighborhood structure provides a wider search space and generates more neighboring solutions than N5, N6 and N7 neighborhood structures, thus increasing the probability of finding the optimal solution. Therefore, N8 neighborhood structure dramatically improves the exploitation ability of TS. Secondly, we further design a neighborhood clipping method for N8 neighborhood structure, which can effectively reduce the size of neighboring solutions and avoid invalid operation movements. The neighborhood clipping method saves the computational time of TS and improves its search efficiency. Finally, N8 neighborhood structure can be easily embedded into any algorithm framework to improve its local search ability. Therefore, N8 neighborhood structure outperforms the previous existing neighborhood structures in both solution quality and computational time.

## 6 Conclusions and future researches

The paper designs a new N8 neighborhood structure and incorporate it into the framework of TS to solve JSP. Afterward, the new neighborhood structure is compared with three state-of-the-art neighborhood structures on four sets of standard benchmark instances. The computational results demonstrate that our neighborhood structure performs better than the othe neighborhood structures in both solution quality and computational time.

The contributions of our research are summarized as follows:

1) A new neighborhood structure is designed for JSP. It is easily incorporated into meta-heuristics to improve their local search ability.
2) A neighborhood clipping method is proposed to avoid invalid insertion and reduce the size of neighborhood solutions.

Although our proposed method can effectively solve JSP, there are still some limitations in this paper. One of these limitations is that although we propose a neighborhood clipping method, the size of neighborhood solutions is still relatively large. This shortcoming increases the computational time of our algorithm. For future work, one research direction is to improve our neighborhood clipping method and make it better to reduce the size of neighborhood solutions. Another interesting research direction is to incorporate our neighborhood structure into other algorithm frameworks to solve JSP.

## Acknowledgments

This work was supported by the National Natural Science Foundation of China under Grant 51825502 and Grant 51721092, and by the Program for HUST Academic Frontier Youth Team under Grant 2017QYTD04.